\documentclass[11pt,a4paper]{article}
\usepackage[hyperref]{naaclhlt2018}

\usepackage{times}
\usepackage{latexsym}

\usepackage{url}

\usepackage{graphicx}
\usepackage{amssymb,amsthm,amsmath}

\DeclareMathOperator*{\argmax}{arg\,max}

\usepackage{colortbl,array,xcolor}
\usepackage{multirow}
\usepackage{makecell}
\usepackage{url}
\usepackage{color}
\usepackage{caption}
\usepackage{subcaption,siunitx,booktabs}
\usepackage{sidecap}
\usepackage{multirow}
\usepackage{textcomp}
\usepackage{url}
\usepackage{pbox}
\usepackage{dblfloatfix}
\usepackage{here}
\usepackage{caption}

\usepackage{enumitem,setspace}
\usepackage[normalem]{ulem}
\usepackage[compact]{titlesec}

\aclfinalcopy 

\newcommand{\veci}[1]{\mbox{\boldmath $#1$}}
\newcommand\scaleeq[3]{\mbox{\scalebox{#1}[#2]{$#3$}}}

\title{Robust Multilingual 
Part-of-Speech Tagging 
via Adversarial Training}
\author{Michihiro Yasunaga \quad\quad\quad Jungo Kasai \quad\quad\quad Dragomir Radev\\
Department of Computer Science, Yale University\\
\scalebox{0.85}[0.9]{{\tt \{michihiro.yasunaga,jungo.kasai,dragomir.radev\}@yale.edu}}}

\date{}

\begin{document}
\setlength{\abovedisplayskip}{4pt}
\setlength{\belowdisplayskip}{4pt}
\setlength{\extrarowheight}{4pt} 

\newcolumntype{G}{>{\columncolor[gray]{0.9}}c} 

\maketitle

\begin{abstract}
Adversarial training (AT)\footnote{We distinguish AT from  Generative Adversarial Networks (GANs).} 
is a powerful regularization method for neural networks, aiming to achieve robustness to input perturbations.
Yet, the specific effects of the robustness obtained from AT are still unclear
in the context of natural language processing.
In this paper, we propose and analyze a neural POS tagging model that exploits AT.
In our experiments on the Penn Treebank WSJ corpus
and the Universal Dependencies (UD) dataset (27 languages),
we find that AT
not only improves the overall tagging accuracy, but also
1) prevents over-fitting well in low resource languages and
2) boosts tagging accuracy for rare \!/\! unseen words.
We also demonstrate that 3) the improved tagging performance by AT contributes to the downstream task of dependency parsing, and that 4)
AT helps the model to learn cleaner
word representations.
5) The proposed AT model is generally effective in different sequence labeling tasks.
These positive results motivate further use of AT for natural language tasks.
\end{abstract}

\section{Introduction}
\begin{figure}[!h]
    \hspace{-3mm}
    \centering
    \includegraphics[width=0.49\textwidth]{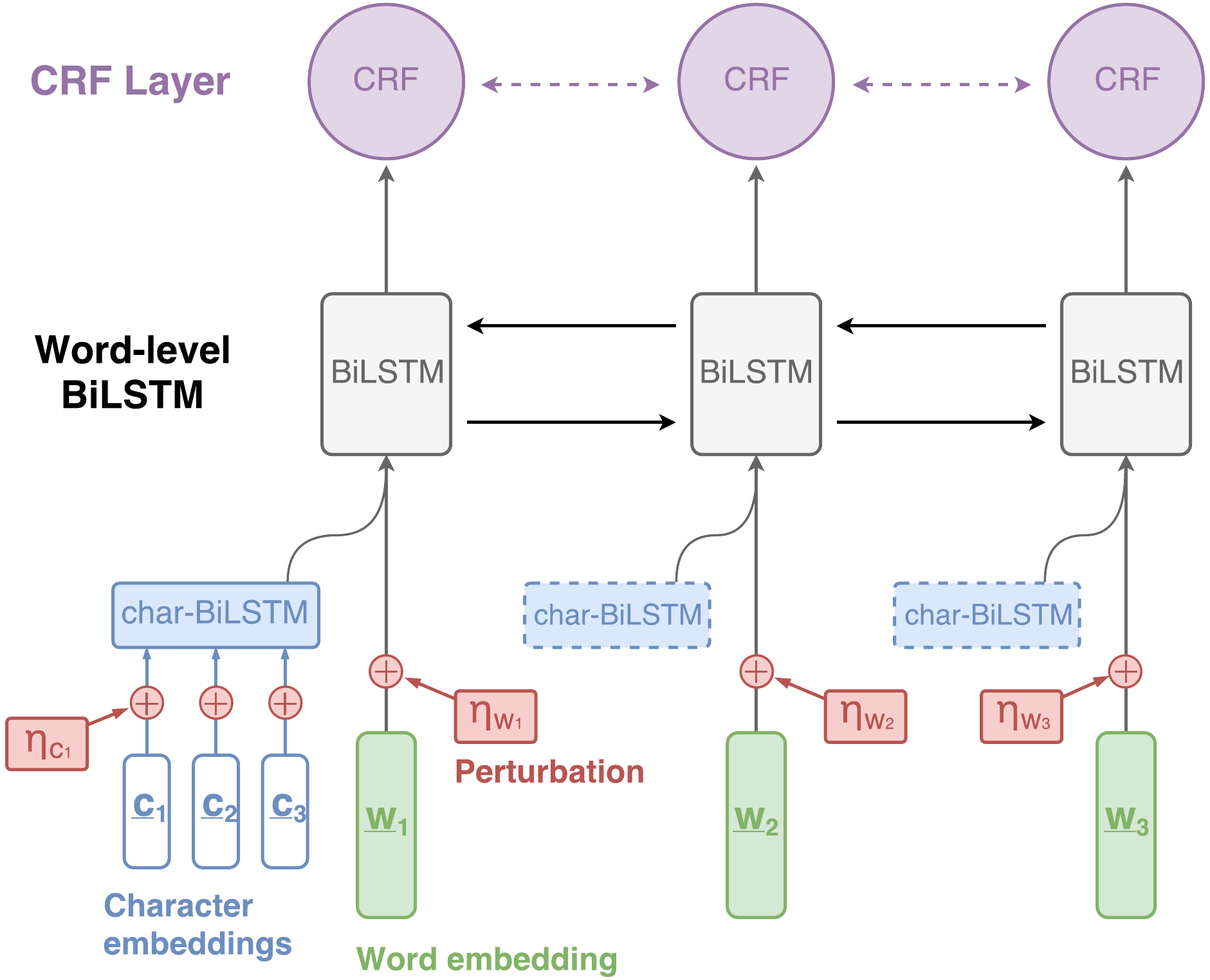}
    \caption{Illustration of our architecture for adversarial POS tagging.
    Given a sentence, we input the normalized word embeddings ($\veci{\uline{w}_1}, \veci{\uline{w}_2}, \veci{\uline{w}_3}$) and character embeddings (showing $\veci{\uline{c}_1}, \veci{\uline{c}_2}, \veci{\uline{c}_3}$ for $w_1$). 
    Each word is represented by concatenating its word embedding and 
    its character-level BiLSTM output.
    They are fed into the main BiLSTM-CRF network for POS tagging.
    In adversarial training, we compute and add the worst-case perturbation $\eta$ to all the input embeddings for regularization.
    }
    \label{fig:adv_pic}
\vspace{-4mm}
\end{figure}

Recently, neural network-based approaches have become popular in many natural language processing (NLP) tasks including tagging, parsing, and translation \cite{chen2014fast,bahdanau2014neural,ma-hovy:2016:P16-1}.
However, it has been shown that neural networks tend to be locally unstable
and
even tiny perturbations to the original inputs can mislead the models \cite{Szegedy2013}. Such maliciously perturbed inputs are called {\it adversarial examples}.
{\it Adversarial training} \cite{Goodfellow2015explain} aims to improve the robustness of a model to input perturbations by training on both unmodified examples and adversarial examples.
Previous work \cite{Goodfellow2015explain,shaham2015understanding} on image recognition 
has demonstrated the enhanced robustness of their models to unseen images via adversarial training and
has provided theoretical explanations of
the regularization effects.

Despite its potential as a powerful regularizer, adversarial training (AT) has yet to be explored extensively in natural language tasks.
Recently, \newcite{miyato2017adv} applied AT on text classification,
achieving state-of-the-art accuracy.
Yet, the specific effects of the robustness obtained from AT are still unclear in the context of NLP.
For example,
research studies have yet to answer questions such as
1) how can we interpret perturbations or robustness on natural language inputs?
2) how are they related to linguistic factors like vocabulary statistics?
3) are the effects of AT language-dependent?
Answering such questions is crucial to understand and motivate the application of adversarial training on natural language tasks.

In this paper,
spotlighting a well-studied core problem of NLP, we propose and carefully analyze a neural part-of-speech (POS) tagging model that exploits adversarial training.
With a BiLSTM-CRF model \cite{Huang2015BidirectionalLM,ma-hovy:2016:P16-1} as our baseline POS tagger, we
apply adversarial training by considering perturbations to input word \!/\! character embeddings.
In order to demystify the effects of adversarial training in the context of NLP, we conduct POS tagging experiments on multiple languages using the Penn Treebank WSJ corpus (Englsih) and the Universal Dependencies dataset (27 languages), with thorough analyses of the following points: \vspace{0mm}
\begin{itemize}[topsep=2mm]
\setlength{\parskip}{-1.4mm}
\setlength{\leftskip}{-0mm}
    \item Effects on different target languages
    \item Vocabulary statistics and tagging accuracy
    \item Influence on downstream tasks
    \item Representation learning of words\vspace{-1mm}
\end{itemize}

In our experiments, we find that our adversarial training model consistently
outperforms the baseline POS tagger, and even achieves state-of-the-art results on 22 languages.
Furthermore,
our analyses reveal the following 
insights into adversarial training in the context of NLP: \vspace{-1mm}

\begin{itemize}[topsep=2mm]
\setlength{\leftskip}{-3mm}
\setlength{\parskip}{0mm}
    \item The regularization effects of  adversarial training (AT) are general across different languages. AT can prevent overfitting especially well when training examples are scarce, providing an effective tool to process low resource languages.

    \item AT can boost the tagging performance for rare \!/\! unseen words and increase the sentence-level accuracy. This positively affects the performance of down-stream tasks such as dependency parsing, where low sentence-level POS accuracy can be a bottleneck \cite{Manning:2011:from97to100}.
    
    \item AT helps the network learn cleaner word embeddings, showing stronger correlations with their POS tags. \vspace{-1mm} 
\end{itemize}
We argue that the effects of AT 
can be interpreted from the perspective of natural language.
Finally, we demonstrate that the proposed AT model is generally effective across different sequence labeling tasks.
This work therefore provides a strong motivation and basis for utilizing adversarial training in
NLP 
tasks.

\section{Related Work}
\subsection{POS Tagging}

Part-of-speech (POS) tagging is a fundamental NLP task that facilitates downstream tasks such as syntactic parsing.
While current state-of-the-art POS taggers \cite{wang:2015,ma-hovy:2016:P16-1} yield accuracy over 97.5\% on PTB-WSJ, there still remain issues.
The per token accuracy metric is easy since taggers can easily assign correct POS tags to highly unambiguous tokens, such as punctuation \cite{Manning:2011:from97to100}.
Sentence-level accuracy serves as a more realistic metric for POS taggers but it still remains low. 
Another problem with current POS taggers is that their accuracy deteriorates drastically on low resource languages 
and rare words \cite{plank2016multilingual}.
In this work, we demonstrate that adversarial training (AT) can mitigate these issues.

It is empirically shown that POS tagging performance can greatly affect downstream tasks such as dependency parsing \cite{dozat-qi-manning:2017:K17-3}. 
In this work, we also demonstrate that the improvements obtained from our AT POS tagger actually contribute to
dependency parsing. 
Nonetheless, parsing with gold POS tags still yields better results, bolstering the view that POS tagging is an essential task in NLP that needs further development.

\subsection{Adversarial Training}
The concept of adversarial training 
\cite{Szegedy2013,Goodfellow2015explain} was originally introduced in the context of image classification to improve the robustness of a model by training on input images with malicious perturbations.
Previous work
\cite{Goodfellow2015explain,shaham2015understanding,wang2016theoretical} has provided a theoretical framework to understand adversarial examples and the regularization effects of adversarial training (AT) in image recognition.

Recently, \newcite{miyato2017adv} applied AT to a natural language task (text classification)
by extending the concept of adversarial perturbations to word embeddings.
\newcite{Wu2017adv} further explored the possibility of AT in relation extraction.
Both report improved performance on their tasks via AT, but
the specific effects of AT have yet to be analyzed.
In our work, we aim to address this issue by providing detailed analyses on the effects of 
AT from the perspective of NLP, such as different languages, vocabulary statistics, word embedding distribution, and aim to motivate future research that exploits AT in NLP tasks.

AT is related to other  regularization methods that add noise to data
such as dropout \cite{JMLR:v15:srivastava14a} and its variant for NLP tasks, word dropout \cite{Iyyer:Manjunatha:Boyd-Graber:III-2015}.
\newcite{xie17_data_noising} discuss various data noising techniques for language modeling.
While these methods produce random noise, AT generates perturbations that the current model is particularly vulnerable to, and thus is claimed to be effective \cite{Goodfellow2015explain}.

It should be noted that while related in name, adversarial training (AT)
differs from Generative Adversarial Networks (GANs) \cite{goodfellow2014generative}. GANs have already been applied to NLP tasks such as dialogue generation \cite{li2017adversarial} and transfer learning \cite{kim2017cross,gui2017part}. Adversarial training also differs from adversarial \textit{evaluation}, recently proposed for reading comprehension tasks \cite{jia2017adversarial}.

\section{Method}
In this section, we introduce our baseline POS tagging model and explain how we implement adversarial training on top.

\subsection{Baseline POS Tagging Model}
Following the recent top-performing models for sequence labeling tasks \cite{plank2016multilingual,Lample2016ner,ma-hovy:2016:P16-1},
we employ a Bi-directional LSTM-CRF model as our baseline (see Figure \ref{fig:adv_pic} for an illustration).

\paragraph{Character-level BiLSTM.}
Prior work has shown that incorporating character-level representations of words can boost POS tagging accuracy by capturing morphological information present in each language.
Major neural character-level models 
include the character-level CNN \cite{ma-hovy:2016:P16-1} and (Bi)LSTM \cite{dozat-qi-manning:2017:K17-3}.
A Bi-directional LSTM (BiLSTM) \cite{Hochreiter:1997:LSM:1246443.1246450,Schuster:1997:BRN:2198065.2205129} processes each sequence both forward and backward to capture sequential information,
while preventing the vanishing \!/\! exploding gradient problem.
We observed that the character-level BiLSTM outperformed the CNN by 0.1\% on the PTB-WSJ development set, and hence in all of our experiments
we use the character-level BiLSTM.
Specifically, we generate a character-level representation for each word
by feeding its character embeddings into the BiLSTM and obtaining the 
concatenated final states.

\paragraph{Word-level BiLSTM.}

Each word in a sentence is represented by concatenating its word embedding and its character-level representation.
They are 
fed into another level of BiLSTM (word-level BiLSTM) to process the entire sentence.

\paragraph{CRF.}
In sequence labeling tasks it is beneficial to consider the correlations between neighboring labels and jointly decode the best chain of labels for a given sentence.
With this motivation, we apply a conditional random field (CRF) \cite{lafferty:2001} on top of the word-level BiLSTM to perform POS tag inference with global normalization, addressing the ``label bias" problem.
Specifically, given an input sentence, we pass the output sequence of the word-level BiLSTM to a first-order chain CRF to compute the conditional probability of the target label sequence: 
$$p(\veci{y} \,|\, \veci{s}; \veci{\theta})$$
where \veci{\theta} represents all of the model parameters (in the BiLSTMs and CRF), \veci{s} and \veci{y} denote the input embeddings and the target POS tag sequence, respectively, for the given sentence. 

For training, we minimize the negative log-likelihood (loss function)
\begin{equation}
\label{eq:baseline_loss}
 L(\veci{\theta}; \veci{s}, \veci{y}) = - \log p(\veci{y} \,|\, \veci{s}; \veci{\theta})
\end{equation}
with respect to the model parameters.
Decoding searches for the POS tag sequence $\veci{y}^*$ with the highest conditional probability using the Viterbi algorithm.
For more detail about the BiLSTM-CRF formulation, refer to \newcite{ma-hovy:2016:P16-1}.

\subsection{Adversarial Training}
\label{sec:method:adv_training}
{\it Adversarial training} \cite{Goodfellow2015explain} is a powerful regularization method, primarily explored in
image recognition to improve the robustness of classifiers to input perturbations. 
Given a classifier, we first generate input examples that are very close to original inputs (so should yield the same labels) yet are likely to be misclassified by the current model.
Specifically, these {\it adversarial examples} 
are generated by adding small perturbations to the inputs in the direction that significantly increases the loss function
of the classifier ({\it worst-case} perturbations).
Then, the classifier is trained on the mixture of clean examples and adversarial examples to improve the stability to input perturbations.
In this work, we
incorporate adversarial training into our baseline POS tagger, aiming to achieve better regularization effects and to provide their interpretations in the context of NLP.

\paragraph{Generating adversarial examples.}

Adversarial training (AT) considers continuous perturbations to inputs, so
we define perturbations at the level of dense word / character embeddings rather than one-hot vector representations, similarly to \newcite{miyato2017adv}.
Specifically, given an input sentence,
we consider the concatenation of all the word \!/\! character embeddings in the sentence: 
$\veci{s} = [\veci{w}_1, \veci{w}_2, \dots, \veci{c}_1, \veci{c}_2, \dots]$.
To prepare an adversarial example, we aim to generate the worst-case perturbation of a small bounded norm $\epsilon$ that maximizes the loss function $L$ of the current model:
\begin{equation*}
\veci{\eta} = \argmax_{
\scalebox{0.75}{$\veci{\eta}'$}:\,
\Vert\scalebox{0.75}{$\veci{\eta}'$}\Vert_2\,
\leq \,\epsilon }
L(\hat{\veci{\theta}}; \veci{s}\!\!~ + \!\!~\veci{\eta}', \veci{y})
\end{equation*}
where $\hat{\veci{\theta}}$ is the current value of the model parameters, treated as a constant, and $\veci{y}$ denotes the target labels.
Since the exact computation of such $\veci{\eta}$ is intractable in complex neural networks, we employ the Fast Gradient Method \cite{liu2017delving,miyato2017adv} i.e. first order approximation to obtain an approximate worst-case perturbation of norm $\epsilon$, by a single gradient computation:
\begin{equation}
\label{eq:perturb}
\veci{\eta} = \epsilon\, \veci{g}/ \Vert\veci{g}\Vert_2
\mbox{, ~where }  \veci{g} = \nabla_{\scalebox{0.8}{$\veci{s}$}} L(\hat{\veci{\theta}}; \veci{s}, \veci{y})
\end{equation}
$\epsilon$ is a hyperparameter to be determined in the development dataset.
Note that the perturbation $\veci{\eta}$ is generated in the direction that significantly increases the loss $L$.
We find such $\veci{\eta}$ against the current model parameterized by $\hat{\veci{\theta}}$, at each training step, and construct an adversarial example by\vspace{-1mm}
$$\veci{s}_{\mathrm{adv}} = \veci{s} + \veci{\eta} \vspace{-1.5mm}$$

However,
if we do not restrict the norm of word \!/\! character embeddings,
the model could trivially learn embeddings of large norms to make the perturbations insignificant.
To prevent this issue, we normalize word \!/\! character embeddings so that they have mean 0 and variance 1 for every entry, as in \newcite{miyato2017adv}.
The normalization is performed  every time we feed input embeddings into the LSTMs and generate adversarial examples.
To ensure a fair comparison, 
we also normalize input embeddings in our baseline model.

While \newcite{miyato2017adv} set the norm of a perturbation $\epsilon$ (Eq \ref{eq:perturb}) to be a fixed value for all input sentences,
to generate adversarial examples for an entire sentence of a variable length and to include character embeddings besides word embeddings, we make the perturbation size $\epsilon$ adaptive to the dimension of the concatenated input embedding $\veci{s} \in \mathbb{R}^D$. 
We set $\epsilon$ to be $\alpha \scaleeq{0.95}{0.95}{\sqrt{D}}$ (i.e., proportional to $\scaleeq{0.95}{0.95}{\sqrt{D}}$), as the expected squared norm of $\veci{s}$ after the embedding normalization 
is $D$.
The scaling factor $\alpha$ is selected from $\{$0.001, 0.005, 0.01, 0.05, 0.1$\}$ 
based on the development performance in each treebank. 
We used 0.01 for PTB-WSJ and UD-Spanish, and 0.05 for the rest.
Note that $\alpha \!=\! 0$ 
would generate no noise (identical to the baseline);
if $\alpha \!=\! 1$, the generated adversarial perturbation would have a norm comparable to the original embedding, which could change the semantics of the input sentence \cite{Wu2017adv}.
Hence, the optimal perturbation scale $\alpha$ should lie in between and be small enough to preserve the semantics of the original input.

\paragraph{Adversarial training.}

At each training step, we
generate adversarial examples against the current model, and train on the mixture of clean examples and adversarial examples to achieve robustness to input perturbations.
To this end, we define the loss function for adversarial training as:
$$ \tilde{L} = \gamma L(\veci{\theta}; \veci{s}, \veci{y}) + (1-\gamma) L(\veci{\theta}; \veci{s}_{\mathrm{adv}}, \veci{y})$$
where
$L(\veci{\theta}; \veci{s}, \veci{y})$, $L(\veci{\theta}; \veci{s}_{\mathrm{adv}}, \veci{y})$ represent the loss from a clean example and the loss from its adversarial example, respectively, and $\gamma $ determines the weighting between them.
We used $\gamma = 0.5$ in all our experiments.
This objective function can be optimized with respect to the model parameters $\veci{\theta}$, in the same manner as the baseline model.

\section{Experiments}
To fully analyze the effects of adversarial training,
we train and evaluate our baseline \!/\! adversarial POS tagging models on both a standard English dataset and a multilingual dataset.

\subsection{Datasets}
As a standard English dataset, we use the Wall Street Journal (WSJ) portion of the Penn Treebank (PTB) \cite{Marcus:1993:BLA:972470.972475}, containing 45 different POS tags.
We adopt the standard split: sections 0-18 for training, 19-21 for development and 22-24 for testing \cite{Collins2002,Manning:2011:from97to100}.

For multilingual POS tagging experiments,
to compare with prior work,
we use treebanks from Universal Dependencies (UD) v1.2 \cite{11234/1-1548} (17 POS) with the given data splits.
We experiment on languages for which pre-trained Polyglot word embeddings \cite{polyglot:2013:ACL-CoNLL} are available, resulting in 27 languages listed in Table \ref{tbl:ud_results}.
We regard languages with less than 60k tokens of training data as low-resource (Table \ref{tbl:ud_results}, bottom), as in \newcite{plank2016multilingual}.

\subsection{Training \& Evaluation Details}

\paragraph{Model settings.}

We initialize word embeddings with 100-dimensional GloVe \cite{pennington-socher-manning:2014:EMNLP2014} for English, and with 64-dimensional Polyglot \cite{polyglot:2013:ACL-CoNLL} for other languages.
We use 30-dimensional character embeddings, and set the state sizes of character \!/\! word-level BiLSTM to be 50, 200  for English, 50, 100 for low resource languages, and 50, 150 for other languages. The model parameters and character embeddings are randomly initialized, 
as in
\newcite{ma-hovy:2016:P16-1}.
We apply dropout \cite{JMLR:v15:srivastava14a} to input embeddings and BiLSTM outputs for both baseline and adversarial training, with dropout rate 0.5.

\paragraph{Optimization.}
We train the model parameters and word \!/\! character embeddings by the mini-batch stochastic gradient descent (SGD) with batch size 10, momentum 0.9, initial learning rate 0.01 and decay rate 0.05. 
We also use a gradient clipping of 5.0 \cite{Pascanu2012}.
The models are trained with
early stopping \cite{caruana2001overfitting} based on the development performance.

\paragraph{Evaluation.}
We evaluate 
per token tagging accuracy on test sets. 
We repeat the experiment three times and report the statistical significance.

\subsection{Results}
\paragraph{PTB-WSJ dataset.}

\begin{table}[!t]
\setlength{\extrarowheight}{1pt}
\centering
\scalebox{0.75}{
\hspace{-1mm}\begin{tabular}{l|c}
\hline
\vrule width 0pt height 11pt depth 4pt {\bf Model} & {\bf Accuracy}\\
\hline
\vrule width 0pt height 11pt
\newcite{N03-1033} & 97.27\\
\newcite{Manning:2011:from97to100} & 97.28\\
\newcite{Collobert:2011:NLP:1953048.2078186}\, & 97.29\\
\newcite{sogaard:2011:ACL-HLT20111} & 97.50\\
\newcite{wang:2015}\, & {\bf 97.78}\\
\newcite{ma-hovy:2016:P16-1} & 97.55\\
\newcite{YangSC17} & 97.55\\
\vrule width 0pt depth 5pt \newcite{HashimotoXTS16} & 97.55\\
\hline
\rowcolor[gray]{.90} \vrule width 0pt height 11pt Ours -- Baseline {\small (BiLSTM-CRF)}~~ & 97.54\\
\rowcolor[gray]{.90} Ours -- Adversarial & 97.58\\
\hline
\end{tabular}}\vspace{-1mm}

\caption{POS tagging accuracy on the PTB-WSJ test set, with other top-performing systems.
}
\label{table:wsj}
\vspace{-4mm}
\end{table}

Table \ref{table:wsj} shows the POS tagging results.
As expected, our baseline (BiLSTM-CRF) model (accuracy 97.54\%) performs on par with other state-of-the-art systems.
Built upon this baseline, our adversarial training (AT) model reaches accuracy 97.58\% thanks to its regularization power, outperforming recent POS taggers except \newcite{wang:2015}.
The improvement over the baseline is statistically significant, with $p$-value \scalebox{0.9}{$<$} 0.05 on the $t$-test.
We provide additional analysis on this result in later sections.

\paragraph{Multilingual dataset (UD).}

\begin{figure*}[!h]
    ~\vspace{-1.4mm}
    \centering
    \includegraphics[width=0.89\textwidth]{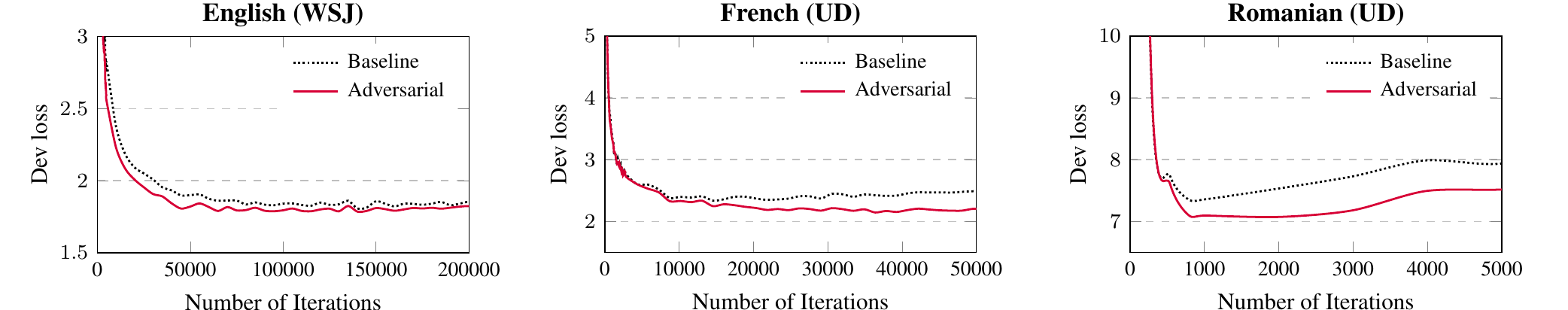}
    \caption{Learning curves for three representative languages (Romanian is low-resource). We show the transition of loss (defined in Eq \protect\ref{eq:baseline_loss}) on the development sets.
    }
    \label{fig:learning_curve}
\vspace{-4mm}
\end{figure*}

\begin{table}[!t]
    \hspace{-2mm}\vspace{-1mm}
    \includegraphics[width=0.495\textwidth]{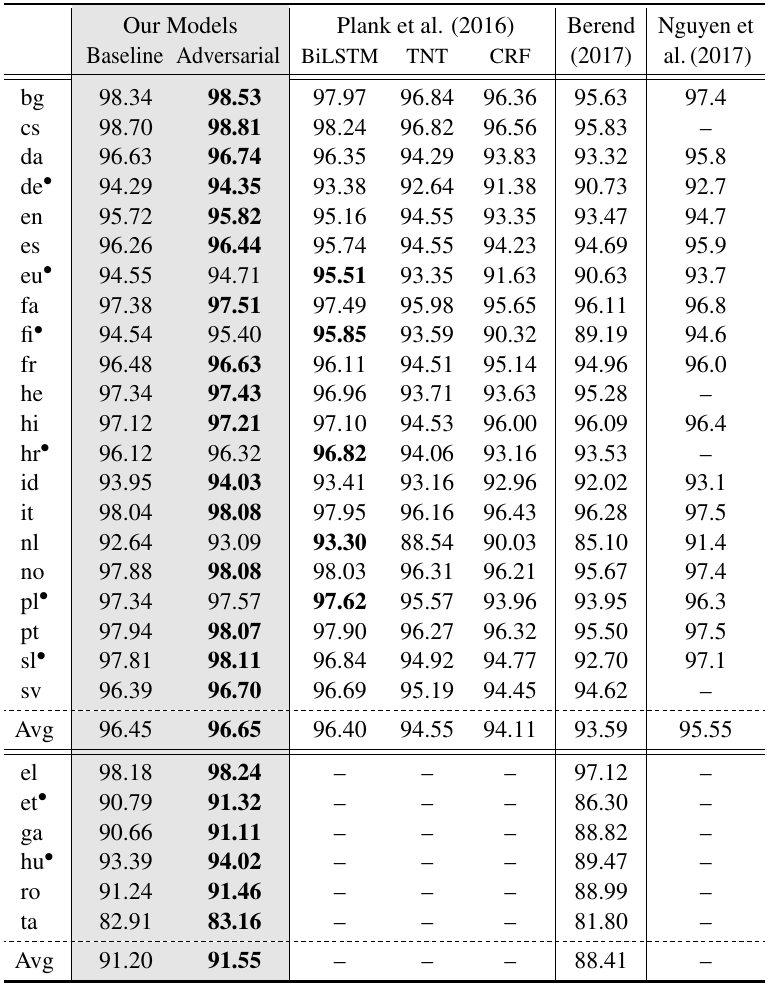}
    \caption{
    POS tagging accuracy (test) for 27 UD v1.2 treebanks, with other recent works,  \protect\newcite{plank2016multilingual}, \protect\newcite{berend2016sparse} and \protect\newcite{Nguyen-etal-2017}.
    For \protect\newcite{plank2016multilingual}, we include the traditional baselines TNT and CRF, and their state-of-the-art model that employs a multi-task BiLSTM.
    Languages with $^\bullet$ are morphologically rich, and 
    those at the bottom (`el' to `ta') are low-resource, containing less than 60k tokens in their training sets.
    }
\label{tbl:ud_results}
\vspace{-4mm}
\end{table}

Experimental results are summarized in Table \ref{tbl:ud_results}.
Our AT model shows clear advantages over the baseline in all of the 27 languages (average improvement $\sim$0.25\%; see the two shaded columns). 
Considering that our baseline (BiLSTM-CRF) is already a top performing model for POS tagging,
these improvements made by AT are substantial.
The improvements are also statistically significant for all the languages, with $p$-value \scalebox{0.9}{$<$} 0.05 on the $t$-test,
suggesting that the regularization by AT 
is generally effective across different languages.
Moreover, our AT model achieves state-of-the-art on nearly all of the languages, except the five where 
\newcite{plank2016multilingual}'s multi-task BiLSTM yielded better results.
Among the five, most languages are morphologically rich ($^\bullet$).\footnote{We followed the criteria of morphological richness used in \newcite{Nguyen-etal-2017}.} 
We suspect that their joint training of word rarity may be of particular help in processing morphologically complex words.

\begin{table}[!t]
    \setlength{\extrarowheight}{1.7pt}
    ~\\[-7.8mm]
    
    \centering
    \begin{flushleft}
    \scalebox{0.75}[0.75]{{\bf English (WSJ)}}
    \end{flushleft}
    \vspace{-1mm}
    
    \scalebox{0.7}{
    \hspace{-1mm}\begin{tabular}{c|cccc|c}
    \Xhline{3\arrayrulewidth}
    \vrule width 0pt depth 5pt \scalebox{1}[1]{Word Frequency}    &  0  & 1-10  &  10-100  &  100- & Total  \\ \Xhline{2\arrayrulewidth}
    \vrule width 0pt height 12pt \# Tokens & 3240 & 7687 & 20908 & 97819 & 129654\\
    Baseline & {\bf 92.25} & 95.36 & 96.03 & 98.19 & 97.53\\
    \vrule width 0pt depth 6pt Adversarial & 92.01 & {\bf 95.52} & \uline{96.10} & \uline{98.23} & \uline{97.57}\\
    \Xhline{3\arrayrulewidth}
    \end{tabular}}
    \vspace{0.5mm}
    
    \centering
    \begin{flushleft}
    \scalebox{0.75}[0.75]{{\bf French (UD)}}
    \end{flushleft}
    \vspace{-1mm}

    \scalebox{0.7}{
    \hspace{-1mm}\begin{tabular}{c|cccc|c}
    \Xhline{3\arrayrulewidth}
    \vrule width 0pt depth 5pt \scalebox{1}[1]{Word Frequency} & 0 & 1-10  &  10-100  &  100- & Total  \\ \Xhline{2\arrayrulewidth}
    \vrule width 0pt height 12pt \# Tokens & 356 & 839 & 1492 & \textcolor{white}{*\hspace{-1mm}}4523\textcolor{white}{*\hspace{-1mm}} & \textcolor{white}{*}7210\textcolor{white}{*}\\
    Baseline & 87.64 & 94.05 & 94.03 & 98.43 & 96.48\\
    \vrule width 0pt depth 6pt Adversarial &  \uline{87.92} & {\bf 94.88} & 94.03 & \uline{98.50} & \uline{96.63}\\
    \Xhline{3\arrayrulewidth}
    \end{tabular}}\\[-1mm]
    
    \caption{POS tagging accuracy (test) on different subsets of words, categorized by their frequency of occurrence in training.
    The second row shows the number of tokens in the test set that are in each category. 
    The third and fourth rows show the performance of our two models.
    Better scores are \uline{underlined}. The biggest improvement is {\bf in bold}.
    }
    \label{table:word_freq}
    \vspace{-4mm}
\end{table}

Additionally, we see that our AT model achieves notably large improvements over the baseline in resource-poor languages (the bottom of Table \ref{tbl:ud_results}), with average improvement 0.35\%, as compared to that for resource-rich languages, 0.20\%.
To further visualize the regularization effects, we present the learning curves for three representative languages, English (WSJ), French (UD-fr) and Romanian (UD-ro, low-resource), based on the development loss (see Figure \ref{fig:learning_curve}).
For all the three languages, we can observe that the AT model (red solid line) prevents overfitting better than the baseline (black dotted line), and this advantage is more significant in low resource languages.
For example, in Romanian, 
the baseline model starts to increase development loss after 1,000 iterations even with dropout, whereas the AT model keeps improving until 2,500 iterations, achieving notably lower development loss (0.4 down).
These results illustrate that AT can prevent overfitting especially well on small datasets and can augment the regularization power beyond dropout.
AT can also be viewed as an effective means of data augmentation, where we generate and train with new examples the current model is particularly vulnerable to at every time step, enhancing the robustness of the model.
AT can therefore be a promising tool to process low resource languages.

\section{Analysis}

\begin{table}[!t]
    \setlength{\extrarowheight}{1.7pt}
    ~\\[-7.8mm]
    
    \centering
    \begin{flushleft}
    \scalebox{0.75}[0.75]{{\bf English (WSJ)}}
    \end{flushleft}
    \vspace{-1mm}
    
    \scalebox{0.7}{
    \hspace{-1mm}\begin{tabular}{c|cccc|c}
    \Xhline{3\arrayrulewidth}
    \vrule width 0pt depth 5pt \scalebox{1}[1]{Word Frequency} &  0  & 1-10  &  10-100  & 100- & Total  \\ \Xhline{2\arrayrulewidth}
    \vrule width 0pt height 12pt \# Tokens & 6480 & 15374 & 41815 & \!195637 & 259306\\
    Baseline & 97.76 & 97.71 & 97.80 & 97.45 & 97.53\\
    \vrule width 0pt depth 6pt Adversarial & {\bf 98.06} & 97.71 & \uline{97.89} & \uline{97.47} & \uline{97.57}\\
\Xhline{3\arrayrulewidth}
    \end{tabular}}
    ~\\[-0.5mm] 
    
    \centering
    \begin{flushleft}
    \scalebox{0.75}[0.75]{{\bf French (UD)}}
    \end{flushleft}
    \vspace{-1mm}
    
    \scalebox{0.7}{
    \hspace{-1mm}\begin{tabular}{c|cccc|c}
    \Xhline{3\arrayrulewidth}
    \vrule width 0pt depth 5pt \scalebox{1}[1]{Word Frequency} & 0 & 1-10  &  10-100  &  100- & Total  \\ \Xhline{2\arrayrulewidth}
    \vrule width 0pt height 12pt \# Tokens & 712 & \textcolor{white}{*\hspace{-1.3mm}}1678\textcolor{white}{*\hspace{-1.3mm}} & 2983 & \textcolor{white}{*}9045\textcolor{white}{*} & \textcolor{white}{*\hspace{-1mm}}14418\textcolor{white}{*\hspace{-1mm}}\\
    Baseline & 95.08 & 97.08 & 97.58 & 96.11 & 96.48\\
    \vrule width 0pt depth 6pt Adversarial & {\bf 95.37} & \uline{97.26} & \uline{97.79} & \uline{96.23} & \uline{96.63}\\
    \Xhline{3\arrayrulewidth}
    \end{tabular}}\\[-1mm]
    
    \caption{POS tagging accuracy (test) on {\it neighboring} words. 
    We cluster all words in the test set in the same way as Table \protect\ref{table:word_freq} and consider the tagging performance on the neighbors (left and right) of these words in the test text.
    }
    \label{table:neighbor}
    \vspace{-4mm}
\end{table}

In the previous sections, we demonstrated the regularization power of adversarial training (AT) on different languages, based on the overall POS tagging performance and learning curves.
In this section, we conduct further analyses on the robustness of AT from NLP specific aspects such as word statistics, sequence modeling, downstream tasks, and word representation learning.

We find that AT can boost tagging accuracy on rare words and neighbors of unseen words (\S \ref{Word-level_Analysis}).
Furthermore, this robustness against rare \!/\! unseen words leads to better sentence-level accuracy and downstream dependency parsing (\S \ref{Sentence-level_Analysis}).
We illustrate these findings using
two major languages, English (WSJ) and French (UD), 
which have substantially large training and testing data to discuss vocabulary statistics and sentence-level performance.
Finally, we study the effects of AT on word representation learning (\S \ref{representation_Analysis}), and {the applicability of AT to different sequential tasks} (\S \ref{sec:additional_task}).

\subsection{Word-level Analysis}
\label{Word-level_Analysis}
Poor tagging accuracy on rare \!/\! unseen words is one of the bottlenecks in current POS taggers \cite{Manning:2011:from97to100,plank2016multilingual}.
Aiming to reveal the effects of 
AT on rare \!/\! unseen words, we analyze tagging performance at the word level, considering vocabulary statistics.

\paragraph{Word frequency.}

To define rare \!/\! unseen words, we consider each word's frequency of occurrence in the training set. 
We categorize all words in the test set based on this frequency and
study the test tagging accuracy for each group (see Table \ref{table:word_freq}).\footnote{
To conduct the analysis, we picked the median result from the three repeated experiments.}
In both languages, the AT model achieves large improvements over the baseline on rare words (e.g., frequency 1-10 in training), as opposed to more frequent words.
This result again corroborates the data augmentation power of AT under small training examples.
On the other hand, we did not observe meaningful improvements
on unseen words (frequency 0 in training). 
A possible explanation is that AT can facilitate the learning of words with at least a few occurrences in training (rare words),
but is not particularly effective in inferring the POS tags of words for which no training examples are given (unseen words).

\paragraph{Neighboring words.}

One important characteristic of natural language tasks is the sequential nature of inputs (i.e., sequence of words), where each word influences the function of its neighboring words.
Since our model uses BiLSTM-CRF for that reason, we also study the tagging performance on the neighbors of rare \!/\! unseen words, and analyze the effects of AT with the sequence model in mind.
In Table \ref{table:neighbor},
we cluster all words in the test set based on their frequency in training again, and consider the tagging accuracy on the neighbors (left and right) of these words in the test text.
We observe that AT tends to achieve large improvements over the baseline on the neighbors of unseen words (training frequency 0),
while the improvements on the neighbors of more frequent words remain moderate.
Our AT model thus exhibits strong stability to uncertain neighbors, as compared to the baseline.
We suspect that because
we generate adversarial examples against entire input sentences,
training with adversarial examples makes the model more robust not only to perturbations in each word but also to perturbations in its neighboring words, 
leading to greater stability to uncertain neighbors.

\subsection{Sentence-level \& Downstream Analysis}
\label{Sentence-level_Analysis}

\begin{table}[!t]
    \setlength{\extrarowheight}{0pt}
    ~\\[-6mm]
    
    \scalebox{0.75}[0.75]{{\bf English (WSJ)}}\\[0.7mm]
    \scalebox{0.67}{
    \hspace{0mm}\begin{tabular}{c||c||cc|cc}
    \Xhline{3\arrayrulewidth}
     \multicolumn{1}{c||}{\vrule width 0pt height 12pt} & \multicolumn{1}{c||}{\raisebox{-0pt}{\scalebox{0.95}[1]{Sentence-}}} & \multicolumn{2}{c|}{ \scalebox{1}[1]{Stanford Parser}} & \multicolumn{2}{c}{\scalebox{0.95}[1]{Parsey McParseface}} \\
     &  \raisebox{1.5pt}{\scalebox{0.95}[1]{level Acc.\!}} & UAS & LAS & ~\,~UAS~~\, & LAS \\ \Xhline{2\arrayrulewidth}
    Baseline \vrule width 0pt height 12pt depth 0pt &  59.08 & 91.53 & 89.30 & 91.68 & 87.92\\
    Adversarial \vrule width 0pt height 10.5pt depth 0pt & {\bf 59.61} & {\bf 91.57} & {\bf 89.35} & {\bf 91.73} & {\bf 87.97}\\
    \scalebox{0.95}[1]{(w/ gold tags)\!} \vrule width 0pt height 10.5pt depth 6pt & -- & (92.07) & (90.63) & (91.98) & (88.60)\\
    \Xhline{3\arrayrulewidth}
    \end{tabular}}
    ~\\[-0.5mm]
    
    \scalebox{0.75}[0.75]{{\bf French (UD)}}\\[0.7mm]
    \scalebox{0.67}{
    \hspace{0mm}\begin{tabular}{c||c||cc}
    \Xhline{3\arrayrulewidth}
     \multicolumn{1}{c||}{\vrule width 0pt height 12pt} & \multicolumn{1}{c||}{\raisebox{-0pt}{\scalebox{0.95}[1]{Sentence-}}} & \multicolumn{2}{c}{ \scalebox{1}[1]{Parsey Universal}} \\
     &  \raisebox{1.5pt}{\scalebox{0.95}[1]{level Acc.\!}} & ~~UAS~~ & LAS  \\ \Xhline{2\arrayrulewidth}
    Baseline \vrule width 0pt height 12pt depth 0pt & 52.35 & 84.85 & 80.36 \\
    Adversarial \vrule width 0pt height 10.5pt depth 0pt & {\bf 53.36} & {\bf 85.01} & {\bf 80.55}\\
    \scalebox{0.95}[1]{(w/ gold tags)\!} \vrule width 0pt height 10.5pt depth 6pt & -- & (85.05) & (80.75)\\
    \Xhline{3\arrayrulewidth}
    \end{tabular}}\\[-1.5mm]
    
    \caption{Sentence-level accuracy and downstream dependency parsing performance by our baseline \!/\! adversarial POS taggers.
    }
    \label{table:sent_level}
    \vspace{-4mm}
\end{table}

In the word-level analysis, we showed that 
AT can boost tagging accuracy on rare words and the neighbors of unseen words, enhancing overall robustness on rare \!/\! unseen words. 
In this section, we discuss the benefit of our improved POS tagger in a major downstream task, dependency parsing.

Most of the recent state-of-the-art dependency parsers take predicted POS tags as input (e.g. \newcite{chen2014fast, andor2016globally, DozatManning17}).
\newcite{dozat-qi-manning:2017:K17-3} empirically show that their
dependency parser gains significant improvements by using POS tags predicted by a Bi-LSTM POS tagger, while POS tags predicted by the UDPipe tagger \cite{Straka2016UDPipeTP} do not contribute to parsing performance as much.
This observation illustrates that POS tagging performance has a great influence on dependency parsing, motivating the hypothesis that the POS tagging improvements gained from our adversarial training help dependency parsing.

To test the hypothesis, we consider three settings in dependency parsing of English and French: using POS tags predicted by the baseline model, using POS tags predicted by the AT model, and using gold POS tags. 
For English (PTB-WSJ), we first convert the treebank into Stanford Dependencies (SD) using Stanford CoreNLP (ver 3.8.0) \cite{manning-EtAl:2014:P14-5}, and then apply two well-known dependency parsers:
Stanford Parser (ver 3.5.0) \cite{chen2014fast} and Parsey McParseface (SyntaxNet) \cite{andor2016globally}.
For French (UD),
we use Parsey Universal from SyntaxNet.
The three parsers are all publicly available and pre-trained on corresponding treebanks.

Table \ref{table:sent_level} shows the results of the experiments. We can observe improvements in both languages by using the POS tags predicted by our AT POS tagger. 
As \newcite{Manning:2011:from97to100} points out, when predicted POS tags are used for downstream dependency parsing, a single bad mistake in a sentence can greatly damage the usefulness of the POS tagger.
The robustness of our AT POS tagger against rare \!/\! unseen words 
helps to mitigate such an issue.
This advantage can also be observed from the AT POS tagger's notably higher sentence-level accuracy than the baseline (see Table \ref{table:sent_level} left).
Nonetheless, gold POS tags still yield better parsing results as compared to the baseline \!/\! AT POS taggers, supporting the claim that POS tagging needs further improvement for downstream tasks.

\subsection{Effects on Representation Learning}
\label{representation_Analysis}

\begin{table}[t]
\setlength{\extrarowheight}{0pt}
~\\[-7.8mm]

\centering
\begin{flushleft}
\scalebox{0.75}[0.75]{{\bf English (WSJ)}}
\end{flushleft}
\vspace{-1mm}

\scalebox{0.7}[0.7]{
\hspace{-0mm}\begin{tabular}{l|cccc|c}
\Xhline{3\arrayrulewidth}
POS Cluster\vrule width 0pt height 11.5pt depth 4.5pt  & {\tt NN} & {\tt VB} & {\tt JJ} & {\tt RB}  & Avg. \\ \Xhline{2\arrayrulewidth}
1) Initial \scalebox{0.8}[0.8]{(GloVe)}\hspace{2.9mm}\vrule width 0pt depth 4pt height 12.5pt  &  0.243	&  0.426 &  0.220 & 0.549  &  0.359\\
2) Baseline \vrule width 0pt depth 4pt height 11pt &  0.280 & 0.431  & {\bf 0.309} &  0.667 & 0.422  \\
3) Adversarial \vrule width 0pt depth 5pt height 11pt & {\bf 0.281} & {\bf 0.436} & {0.306} & {\bf 0.675} & {\bf 0.424}\\
\Xhline{3\arrayrulewidth}
\end{tabular}}
\vspace{0.5mm}

\centering
\begin{flushleft}
\scalebox{0.75}[0.75]{{\bf French (UD)}}
\end{flushleft}
\vspace{-1mm}

\scalebox{0.7}[0.7]{
\hspace{-0mm}\begin{tabular}{l|cccc|c}
\Xhline{3\arrayrulewidth}
POS Cluster\vrule width 0pt height 11.5pt depth 4.5pt  & {\tt NOUN} & {\tt VERB} & {\tt ADJ} & {\tt ADV}  & Avg. \\ \Xhline{2\arrayrulewidth}
1) Initial \scalebox{0.8}[0.8]{(polyglot)\,}\vrule width 0pt depth 4pt height 12.5pt  &  0.215	& 0.233 & 0.210 & 0.540 & 0.299 \\
2) Baseline \vrule width 0pt depth 4pt height 11pt & 0.258 & 0.271  & 0.262 & 0.701 &  0.373\\
3) Adversarial \vrule width 0pt depth 5pt height 11pt &  {\bf 0.263} & {\bf 0.272}  & {\bf 0.263} &  {\bf 0.720} &  {\bf 0.379}\\
\Xhline{3\arrayrulewidth}
\end{tabular}}\\[-1mm]
\caption{
Cluster tightness evaluation for word embeddings, based on the cosine similarity measure.
Higher scores
indicate better clustering (cleaner word vector distribution).
Each row corresponds to word vectors 1) at the beginning, 2) after baseline training, and 3) after adversarial training.
}
\label{tbl:cluster_tightness}
\vspace{-0mm}
\end{table}

\begin{table}[t]
\setlength{\extrarowheight}{0pt}

\scalebox{0.75}[0.75]{{\bf English (WSJ)}}\\[-3.5mm]

\hspace{-1mm}\scalebox{0.67}[0.7]{
\hspace{-0mm}\begin{tabular}{l||cccccc}
\Xhline{3\arrayrulewidth}
\scalebox{1}[1]{Perturbation scale $\alpha$} \vrule width 0pt height 13pt depth 6pt  & 0 & 0.001 & 0.01 & 0.05  & 0.1 & 0.5\\ \Xhline{1\arrayrulewidth}
\scalebox{1}[1]{Avg. cluster tightness}\vrule width 0pt depth 6pt height 13pt & 0.422 & 0.423 & 0.424 & 0.429 & {\bf 0.436} & 0.429\\
\Xhline{3\arrayrulewidth}
\end{tabular}}\\[-2mm]

\caption{Average cluster tightness for word embeddings 
trained with varied perturbation scale $\alpha$ (0 indicates baseline training). }
\label{tbl:cluster_tightness_epsilon}
\vspace{-4mm}
\end{table}

Next, we perform an analysis on representation learning of words (word embeddings) for the English (PTB-WSJ) and French (UD) experiments.
We hypothesize that adversarial training (AT) helps to learn better word embeddings so that the POS tag prediction of a word cannot be influenced by a small perturbation in the input embedding.

To verify this hypothesis, 
we cluster all words in the test set based on their correct POS tags\footnote{We excluded words with multiple tags in the test text.}
and evaluate the tightness of the word vector distribution within each cluster. 
We compare this clustering quality among the three settings: 1) beginning (initialized with GloVe or Polyglot), 2) after baseline training (50 epochs), and 3) after adversarial training (50 epochs), to study the effects of AT on word representation learning.

For evaluating
the tightness of word vector distribution, we employ the
cosine similarity metric, which is widely used as a measure of the closeness between two word vectors
(e.g., \newcite{mikolov2013distributed}; \newcite{pennington-socher-manning:2014:EMNLP2014}).
To measure the tightness of each cluster, we compute the cosine similarity for every pair of words within, and then take the average.
We also report the average tightness across all the clusters.

The evaluation results are summarized in Table \ref{tbl:cluster_tightness}.
We report the tightness scores for the four major clusters: {\it noun}, {\it verb}, {\it adjective}, and {\it adverb} (from left to right).
As can be seen from the table, for both languages, adversarial training (AT) results in cleaner word embedding distributions than the baseline, with a higher cosine similarity within each POS cluster, and with a clear advantage in the average tightness across all the clusters.
In other words, the learned word vectors show stronger correlations with their POS tags.
This result confirms that training with adversarial examples can help to learn cleaner word embeddings so that the meaning \!/\! grammatical function of a word cannot be altered by a small perturbation in its embedding.
This analysis provides a means to
interpret the robustness to input perturbations, from the perspective of NLP.

\paragraph{Relation with perturbation size $\boldsymbol{\epsilon}$.
}

We also study how the size of added perturbations influences word representation learning in adversarial training.
Recall that we set the norm of a perturbation $\epsilon$ 
to be $\alpha \scaleeq{0.95}{0.95}{\sqrt{D}}$, where $D$ is the dimension of the concatenated input embeddings (see \S \ref{sec:method:adv_training}).
For instance, $\alpha \!=\! 0$ 
would produce no noise; $\alpha \!=\! 1$ would generate a perturbation of a norm equivalent to the original word embeddings.
We hypothesize that AT facilitates word representation learning when $\alpha$ is small enough to preserve the semantics of input words, but can hinder the learning when $\alpha$ is too large.
To test the hypothesis, we repeat the clustering evaluation for 
word embeddings trained with varied perturbation scale $\alpha$: 0, 0.001, 0.01, 0.05, 0.1, 0.5 (see Table \ref{tbl:cluster_tightness_epsilon}).
We observe that the quality of 
learned word embedding distribution keeps improving as $\alpha$ goes up from 0 to 0.1, but starts to drop around $\alpha \!=\! 0.5$.
We also find that this optimal $\alpha$ in word embedding learning (i.e., 0.1)
is larger than the $\alpha$ which yielded the best tagging performance on development sets (i.e., 0.01 or 0.05).
A possible explanation is that
while word embeddings
can adapt to relatively large $\alpha$ (e.g., 0.1) during training, as adversarial perturbations are generated at the embedding level, such $\alpha$ could change the semantics of the input from the current tagging model's perspective and hinder the training of {\it tagging}.

\subsection{Other Sequence Labeling Tasks}
\label{sec:additional_task}

Finally, to further confirm the applicability of AT, we experiment with our BiLSTM-CRF AT model in different sequence labeling tasks: chunking and named entity recognition (NER).\vspace{0.5mm}

\begin{table}[!t]
\setlength{\extrarowheight}{1pt}
\centering
\scalebox{0.75}{
\hspace{-1mm}\begin{tabular}{l|c}
\hline
\vrule width 0pt height 11pt depth 4pt {\bf Model} & {\bf F1}\\
\hline
\vrule width 0pt height 11pt
\newcite{tsuruoka2011learning} & {93.81}\\
\newcite{Collobert:2011:NLP:1953048.2078186}\, & {94.32}\\
\newcite{YangSC17} & {94.66}\\
\newcite{suzuki2008semi} & {95.15}\\
\newcite{sogaard2016deep} & {95.56}\\
\newcite{HashimotoXTS16} & {95.77}\\
\vrule width 0pt depth 5pt \newcite{peters2017semi} & {\bf 96.37}\\
\hline
\rowcolor[gray]{.90} \vrule width 0pt height 11pt Ours -- Baseline {\small (BiLSTM-CRF)}~~ & 95.18\\
\rowcolor[gray]{.90} Ours -- Adversarial & 95.25\\
\hline
\end{tabular}}\vspace{-1mm}

\caption{Chunking F1 scores on the CoNLL-2000 task, with other top performing models.}
\label{table:chunk}
\vspace{-2mm}
\end{table}

\noindent{\bf Chunking} can be performed as a sequence labeling task that assigns a chunking tag (B-NP, I-VP, etc.) to each word.
We conduct experiments on the CoNLL 2000 shared task with the standard data split:
PTB-WSJ Sections 15-18 for training and 20 for testing.
We use Section 19 as the development set and employ the IOBES tagging scheme, following \newcite{HashimotoXTS16}.\vspace{0.5mm}

\begin{table}[!t]
\setlength{\extrarowheight}{1pt}
\centering
\scalebox{0.75}{
\hspace{-1mm}\begin{tabular}{l|c}
\hline
\vrule width 0pt height 11pt depth 4pt {\bf Model} & {\bf F1}\\
\hline
\vrule width 0pt height 11pt
\newcite{Collobert:2011:NLP:1953048.2078186}\, & {89.59}\\
\newcite{Huang2015BidirectionalLM}\, & {90.10}\\
\newcite{chiu2016named}\, & {90.91}\\
\newcite{Lample2016ner}\, & {90.94}\\
\newcite{luo2015joint} & 91.20\\
\newcite{ma-hovy:2016:P16-1} & 91.21\\
\vrule width 0pt depth 5pt \newcite{peters2017semi} & {\bf 91.93}\\
\hline
\rowcolor[gray]{.90} \vrule width 0pt height 11pt Ours -- Baseline {\small (BiLSTM-CRF)}~~ & 91.22\\
\rowcolor[gray]{.90} Ours -- Adversarial & 91.56\\
\hline
\end{tabular}}\vspace{-1mm}

\caption{NER F1 scores on the CoNLL-2003 (English) task, with other top performing models.}
\label{table:ner}
\vspace{-4mm}
\end{table}

\noindent{\bf NER} aims to assign an entity type to each word, such as {\it person}, {\it location}, {\it organization}, and {\it misc}.
We conduct experiments on
the CoNLL-2003 (English) shared task \cite{tjong2003introduction}, adopting the IOBES tagging scheme as in \cite{Lample2016ner,ma-hovy:2016:P16-1}.\vspace{0.5mm}

The results are summarized in Table \ref{table:chunk} and \ref{table:ner}.
AT enhanced F1 score from the baseline BiLSTM-CRF model's 95.18 to 95.25 for chunking, and from 91.22 to 91.56 for NER, also significantly outperforming \newcite{ma-hovy:2016:P16-1}.
These improvements made by AT are bigger than that for English POS tagging, most likely due to the larger room for improvement in chunking and NER. 
The improvements are again statistically significant, with $p$-value \scalebox{0.9}{$<$} 0.05 on the $t$-test. 
The experimental results suggest that the proposed adversarial training scheme is generally effective across different sequence labeling tasks. 

Our BiLSTM-CRF AT model did not reach the performance by \newcite{HashimotoXTS16}'s multi-task model and
\newcite{peters2017semi}'s 
state-of-the-art system that incorporates pretrained language models. 
It would be interesting future work to combine the strengths of these joint models (e.g., syntactic and semantic aids) and adversarial training (e.g., robustness).

\section{Conclusion}
We proposed and carefully analyzed a POS tagging model that exploits adversarial training (AT).
In our multilingual experiments, we find that AT achieves substantial improvements on all the languages tested, especially on low resource ones.
AT also enhances the robustness to rare \!/\! unseen words and sentence-level accuracy, alleviating the major issues of current POS taggers, and contributing to the downstream task, dependency parsing.
Furthermore, our analyses on different languages, word \!/\! neighbor statistics and word representation learning reveal the effects of AT from the perspective of NLP.
The proposed AT model is applicable to general sequence labeling tasks.
This work therefore 
provides a strong basis and motivation for utilizing AT in natural language tasks.

\section*{Acknowledgements}

We would like to thank Rui Zhang, Jonathan Kummerfeld, Yutaro Yamada, as well as all the anonymous reviewers for their helpful feedback and suggestions on this work.

\clearpage
\bibliography{naaclhlt2018}
\bibliographystyle{acl_natbib}

\end{document}